\documentclass[twoside,11pt]{article}

\usepackage{jmlr2e}
\usepackage{enumitem}
\usepackage[utf8]{inputenc}

\DeclareFixedFont{\ttb}{T1}{txtt}{bx}{n}{12} 
\DeclareFixedFont{\ttm}{T1}{txtt}{m}{n}{12}  

\usepackage{color}
\definecolor{deepblue}{rgb}{0,0,0.5}
\definecolor{deepred}{rgb}{0.6,0,0}
\definecolor{deepgreen}{rgb}{0,0.5,0}

\usepackage{listings}

\newcommand\pythonstyle{\lstset{
language=Python,
basicstyle=\small,
otherkeywords={self},             
keywordstyle=\ttb\color{deepblue},
emph={MyClass,__init__},          
emphstyle=\ttb\color{deepred},    
stringstyle=\color{deepgreen},
frame=tb,                         
showstringspaces=false            %
}}

\lstnewenvironment{python}[1][]
{
\pythonstyle
\lstset{#1}
}
{}

\newcommand\pythoninline[1]{{\pythonstyle\lstinline!#1!}}

\jmlrheading{1}{2018}{1-48}{4/00}{10/00}{Goyal et. al.}

\ShortHeadings{DynamicGEM: A Library for Dynamic Graph Embedding Methods}{Goyal et. al.}
\firstpageno{1}

\begin{document}

\title{DynamicGEM: A Library for Dynamic Graph Embedding Methods}

\author{\name Palash Goyal\thanks{Sujit Rokka Chhetri and Palash Goyal contributed equally to this article},  Ninareh Mehrabi,\\ Emilio Ferrara \email \{palashgo,ninarehm,emiliofe\}@usc.edu \\
       \addr Department of Computer Science\\
       University of Southern California\\
       Los Angeles, CA 90007, USA
       \AND
       \name Sujit Rokka Chhetri* \email schhetri@uci.edu \\
       \addr Department of Electrical Engineering and Computer Science\\
       University of California-Irvine\\
       Irvine, CA 92697, USA
       \AND
      \name Arquimedes Canedo \email arquimedes.canedo@siemens.com \\
       \addr Siemens Corporate Technology\\
       Princeton, NJ 08540, USA
       }
       

\editor{Leslie Pack Kaelbling}

\maketitle

\begin{abstract}
DynamicGEM is an open-source Python library for learning node representations of dynamic graphs. It consists of state-of-the-art algorithms for defining embeddings of nodes whose connections evolve over time. The library also contains the evaluation framework for four downstream tasks on the network: graph reconstruction, static and temporal link prediction, node classification, and temporal visualization. We have implemented various metrics to evaluate the state-of-the-art methods, and examples of evolving networks from various domains. We have easy-to-use functions to call and evaluate the methods and have extensive usage documentation. Furthermore, DynamicGEM provides a template to add new algorithms with ease to facilitate further research on the topic.
\end{abstract}


\section{Introduction}

Graph embedding methods aim to represent each node of a graph in a low-dimensional vector space while preserving certain graph's properties \citep{goyal2018graph}. Such  methods  have been used to tackle many real-world tasks, e.g.,  friend recommendation in social networks, genome classification in biology networks, and visualizing topics in research using collaboration networks.

More recently, much attention has been devoted to extending static embedding techniques to capture graph evolution. Applications include  temporal link prediction, and understanding the evolution dynamics of network communities. Most methods aim to efficiently update the embedding of the graph at each time step using information from previous embedding and from changes in the graph. Some methods also capture the temporal patterns of the evolution in the learned embedding, leading to improved link prediction performance.

In this paper, we present DynamicGEM, an easy-to-use toolkit of state-of-the-art dynamic graph embedding  methods. In contrast to existing popular graph embedding libraries like GEM~\citep{goyal3gem}, which implement static graph embedding methods, DynamicGEM implements methods which can handle the evolution of networks over time. Further, we provide a comprehensive framework to evaluate the methods by providing support for four tasks on dynamic networks: graph reconstruction, static and temporal link prediction, node classification, and temporal visualization. For each task, our framework includes multiple evaluation metrics to quantify the performance of the methods. We further share synthetic and real networks for evaluation. Thus, our library is an end-to-end framework to experiment with dynamic graph embedding.  The software is available at \url{https://github.com/palash1992/DynamicGEM}. We provide extensive documentation for installation and testing of various methods available in the library. The documentation also contains an easy-to-use guide for quickly adding  new methods to the framework.

\section{Dynamic Graph Embedding Algorithms}
Dynamic graph embedding algorithms aim to capture the dynamics of the network and its evolution. These methods are useful to predict the future behavior of the network, such as future connections within a network. The problem can be defined formally as follows.

Consider a weighted graph $G(V, E)$, with $V$ and $E$ as the set of vertices and edges respectively.
Given an evolution of graph $\mathcal{G} = \lbrace G_1, .., G_T\rbrace$, where $G_t$ represents the state of graph at time $t$, a dynamic graph embedding method aims to represent each node $v$ in a series of low-dimensional vector space $y_{v_1}, \ldots y_{v_t}$ by learning mappings $f_t: \{V_1, \ldots, V_t, E_1, \ldots E_t\} \rightarrow \mathbb{R}^d$ and $y_{v_i} = f_i(v_1, \ldots, v_i, E_1, \ldots E_i)$.
The methods differ in the definition of $f_t$ and the properties of the network preserved by $f_t$.

There are various existing state of the art methods trying to solve this problem that we have incorporated and included them in this python package including: 

\begin{enumerate}
[noitemsep,topsep=0pt,parsep=0pt,partopsep=0pt,leftmargin=*]
\item \textbf{Optimal SVD}: This method decomposes adjacency matrix of the graph at each time step using Singular Value Decomposition (SVD) to represent each node using the $d$ largest singular values~\citep{ou2016asymmetric}.
\item \textbf{Incremental SVD}: This method utilizes a perturbation matrix capturing the dynamics of the graphs along with performing additive modification on the SVD~\citep{BRAND200620}.
\item \textbf{Rerun SVD}: This method uses incremental SVD to create the dynamic graph embedding. In addition to that, it uses a tolerance threshold to restart the optimal SVD calculations and avoid deviation in incremental graph embedding~\citep{zhang2017timers}.
\item \textbf{Dynamic TRIAD}: This method utilizes the triadic closure process to generate a graph embedding that preserves structural and evolution patterns of the graph~\citep{zhou2018dynamic}.
\item \textbf{AEalign}: This method uses deep auto-encoder to embed each node in the graph and aligns the embeddings at different time steps using a rotation matrix~\citep{goyal2018dyngem}.
\item \textbf{dynGEM}: This method utilizes deep auto-encoders to incrementally generate embedding of a dynamic graph at snapshot $t$~\citep{goyal2018dyngem}.
\item \textbf{dyngraph2vecAE}: This method models the interconnection of nodes within and across time using multiple fully connected layers~\citep{goyal2018dyngraph2vec}
\item \textbf{dyngraph2vecRNN}: This method uses sparsely connected Long Short Term Memory (LSTM) networks to learn the embedding \citep{goyal2018dyngraph2vec}
\item \textbf{dyngraph2vecAERNN}: This method uses a fully connected encoder to initially acquire low dimensional hidden representation and feeds this representation into LSTMs to capture network dynamics~\citep{goyal2018dyngraph2vec}.
\end{enumerate}

\section{Software Package}
DynamicGEM implements the state-of-the-art dynamic graph embedding methods in Python and MATLAB and provides a Python interface for all the methods. The library takes as input a series of dynamic graphs as list of Networkx Digraph each corresponding to a single time step. In this section, we discuss details of library usage.

\subsection{Usage}
Next, we demonstrate the visualization of the graph embedding performed by dyngraph2vecRNN on stochastic block model dataset. 

\begin{python}

from time import time
#import helper libraries
from dynamicgem.visualization import plot_dynamic_sbm_embedding as pltdyn
from dynamicgem.graph_generation import dynamic_SBM_graph as sbm
#import the methods
from dynamicgem.embedding.dynRNN       import DynRNN

# Parameters for Stochastic block model graph
node_num      = 1000 #1000 total nodes
community_num = 2 # two communities
node_change_num = 10 #migrate 10 nodes in each time steps
length          = 4 # total time steps

#Generate the dynamic graph
dynamic_sbm_series = list(sbm.get_community_diminish_series_v2(node_num, 
community_num, length,  1, node_change_num))
graphs     = [g[0] for g in dynamic_sbm_series]
#Initialize embedding algorithm
embedding= DynRNN(d=128, beta=5, n_prev_graphs=2, nu1=1e-6, nu2=1e-6,
n_enc_units=[500,300], n_dec_units=[500,300], rho=0.3,n_iter=250, xeta=1e-3, 
n_batch=100, modelfile= ['./intm/enc_model.json', './intm/dec_model.json'],
weightfile=['./intm/enc_wghts.hdf5', './intm/dec_weghts.hdf5'], 
savefilesuffix = "testing"  )
embs = []
for temp_var in range(lookback+1, length+1):
                emb, _ = embedding.learn_embeddings(graphs[:temp_var])
                embs.append(emb)
pltdyn.plot_dynamic_sbm_embedding_v2(embs[-5:-1], dynamic_sbm_series[-5:])    
\end{python}

\begin{figure}[h]
    \centering
    
    \includegraphics[width=0.6\textwidth]{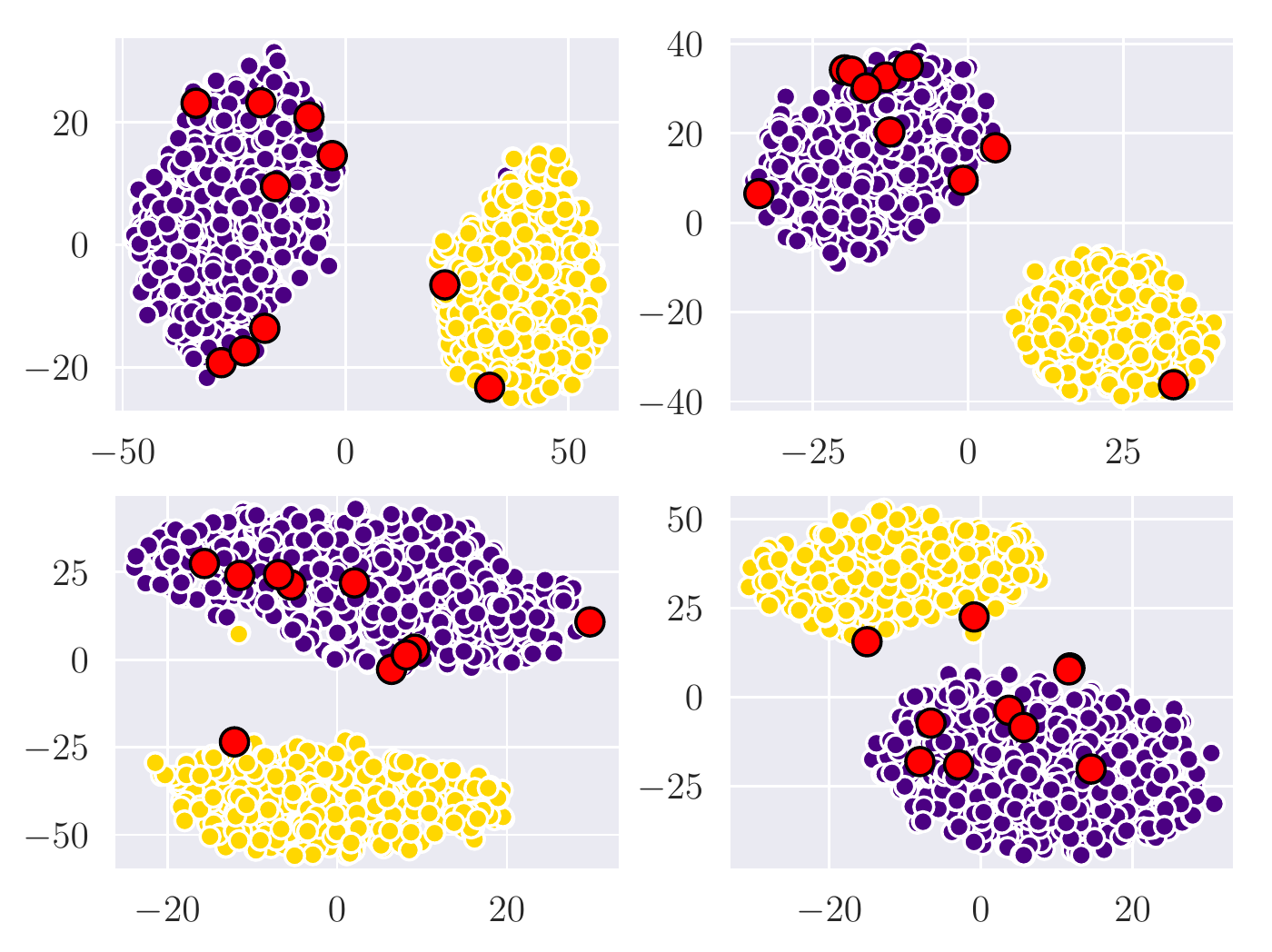}
    \vspace{-1.5em}
    \caption{Visualization of the embedding generated by \textit{dyngraph2vecRNN} after performing dimensionality reduction using TSNE on stochastic block model diminishing community dataset. The red colored nodes migrate from yellow to blue community in next iteration, however based on the temporal learning, \textit{dyngraph2vecRNN} is able embed them closer to the blue community. }
    \vspace{-2em}
    \label{fig:my_label}
\end{figure}

\subsection{Documentation and Design}
DynamicGEM package contains README files in DynamicGEM and its sub directories including dynamicgem/dynamictriad, and dynamicgem/graph-generation directories containing explanation about the repository, its structure, setup, implemented methods, usage, dependencies, and other useful information for user guidance. The repository is organized in an easy to navigate manner. The subdirectories are organized based on the functionality which they serve as follows:
\begin{itemize}
[noitemsep,topsep=0pt,parsep=0pt,partopsep=0pt,leftmargin=*]
\item DynamicGEM/embedding: It contains implementation of the algorithms listed in section~\ref{sec3}. In addition to dynamic graph embedding algorithms, this sub directory contains implementation for some static graph embedding methods on which the dynamic methods are built.
\item DynamicGEM/evaluation: It contains implementations of graph reconstruction, static and temporal link prediction and visualization for evaluation purposes.
\item DynamicGEM/utils: It contains implementation of utility functions for data preparation, plotting, embedding formatting, evaluation, and a variety of other functions that are building blocks of other functions.
\item DynamicGEM/graph-generation: It consists of functions to generate dynamic Stochastic Block Models (SBM)~\citep{wang1987stochastic} with diminishing community.
\item DynamicGEM/visualization: It contains functions for visualizing dynamic and static graph embeddings.
\item DynamicGEM/experiments: It contains useful hyper-parameter tuning function implementations.
\item DynamicGEM/TIMERS: It contains the TIMERS source code.
\item DynamicGEM/dynamicTriad: It contains the dynamicTriad source code.
\end{itemize} \label{sec3}

\section{Conclusion}
DynamicGEM is a Python library with an extensive documentation that includes implementations for a variety of state-of-the-art dynamic graph embedding methods. This library can help researchers and developers to perform a wide range of experiments related to dynamic graphs that evolve over time, their different characteristics, and prediction of these characteristics that can help improvements in this field of study. The library allows easy integration of novel approaches in the domain and functions to evaluate their efficacy. The aim of the library is to evaluate such methods with greater ease.


\acks{The authors are grateful to the Defense Advanced Research Projects Agency  (DARPA), contract W911NF-17-C-0094, for their support. 
We would also like to acknowledge support 
from the National Science Foundation (NSF grant IIS-9988642)
and the Multidisciplinary Research Program of the Department
of Defense (MURI N00014-00-1-0637). }


\vskip 0.2in
\bibliography{sample}

\end{document}